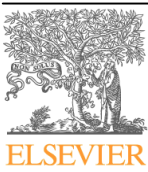

# Hyperspectral Images Classification Using Energy Profiles of Spatial and Spectral Features

Hamid Reza Shahdoosti[a] *

[a] *Hamedan University of Technology, Department of Electrical Engineering, Hamedan, 6515773619, Iran,*

A B S T R A C T

2012 Elsevier Ltd. All rights reserved.

This paper proposes a spatial feature extraction method based on energy of the features for classification of the hyperspectral data. A proposed orthogonal filter set extracts spatial features with maximum energy from the principal components and then, a profile is constructed based on these features. The important characteristic of the proposed approach is that the filter set's coefficients are extracted from statistical properties of data, thus they are more consistent with the type and texture of the remotely sensed images compared with those of other filters such as Gabor. To assess the performance of the proposed feature extraction method, the extracted features are fed into a support vector machine (SVM) classifier. Experiments on the widely used hyperspectral images namely, Indian Pines, and Salinas data sets reveal that the proposed approach improves the classification results in comparison with some recent spectral–spatial classification methods.

## 1. Introduction

The availability of remotely sensed hyperspectral images acquired in the adjacent bands of the electromagnetic spectrum, makes it necessary to propose techniques which are able to interpret such high-dimensional data in many various applications. One of the most important applications in the field of hyperspectral images is the classification in which land covers are distinguished from each other. For this purpose, several techniques have been developed to greatly benefit from the wealth of spectral and spatial information buried in the hyperspectral images.

Various experiments have proven the sparseness of high-dimensional data spaces such that the data structure exists primarily in a subspace (Shahdoosti and Mirzapour, 2017). As a consequence, there is a need for feature extraction methods by which the dimensionality of the data can be reduced to the right subspace without losing the essential information that allows for the separation of classes. These techniques can be mainly classified into two categories (Shahdoosti and Javaheri 2019): 1) Several methods use linear transforms to extract the spectral or spatial information from hyperspectral data. Widely used linear feature extraction methods in the spectral domain include principal component analysis (PCA) (Shahdoosti and Ghassemain, 2012, 2015, 2016), independent component analysis (ICA) (Bayliss et al., 1997) and linear discriminant analysis (Bandos et al., 2009), and those in the spatial domain include the Gabor filter bank (Mirzapour and Ghassemian, 2015) and wavelets. 2) Several techniques exploit spectral or spatial features obtained through nonlinear transformations. Examples of these methods are morphological analysis (Plaza et al., 2005; Benediktsson et al., 2005), kernel methods (Camps-Valls and Bruzzone, 2005), and manifold regularization (Ma et al., 2010).

This paper focuses on using linear transformations for feature extraction methods. Linear transformations are attractive for image processing owing to the fact that they have a very low computational burden while, at the same time they are successful in extracting relevant information from the hyperspectral images.

On the other hand, feature extraction methods can be classified into either supervised or unsupervised ones (Ghassemian and Landgrebe, 1988; Khayat et al. 2008, Shahdoosti and Javaheri 2018). For instance, PCA, Gabor filter bank, ICA and morphological analysis are unsupervised, i.e. these methods do not use the training samples, available for the classification procedure. But, LDA is an example of the supervised methods whose features are extracted by making use of the training samples.

Spectral-based per-pixel classification methods cannot take into account the spatial relationship between the pixels satisfactorily, and it has been widely accepted that the buried spatial information should be exploited as a complementary feature source for distinguishing the land covers. To this end, filter banks are usually used to extract the spatial features of the hyperspectral data, to be used in the classification procedure. A relevant criterion for determining the filter set's coefficients is energy, such that the filters maximizing the energy of features are the appropriate filters for which we look. For this purpose, this paper proposes a directional filter set which extracts spatial information with maximum energy from the remotely sensed images. It will be reasoned that the proposed method has a low computational burden too.

The paper is structured in five sections. Designing the filter set is presented in section 2. In section 3, the energy profile is constructed including the spectral-spatial features of the hyperspectral data. Section 4 presents the experimental results and discussions. Eventually, section 5 is devoted to the conclusions.

## 2. Proposed filter set

The aim is to design a two dimensional directional filter $\mathbf{f} \in \mathrm{R}^{c \times c}$:

$$\mathbf{f} = \begin{bmatrix} f_{1,1} & \cdots & f_{1,c} \\ \cdots & f_{\frac{c+1}{2}, \frac{c+1}{2}} & \cdots \\ f_{c,1} & \cdots & f_{c,c} \end{bmatrix} \quad (1)$$

where $c$ denotes the number of rows and columns of the filter, by which the features of an image $\mathbf{m} \in \mathrm{R}^{r \times l}$ are extracted, where $r$ and $l$ are the number of rows and columns in the image, respectively. By a matrix multiplication, one can describe the convolution operation (Jayaraman et al., 2011):

$$\mathbf{A} = \mathbf{f} * \mathbf{m} = \text{Matrix}\{\mathbf{F}^T \times \mathbf{C}\} \quad (2)$$

where the matrix $\mathbf{A} \in \mathrm{R}^{r \times l}$ is the extracted feature by the filter $\mathbf{f}$ which should be used in the classification procedure, * is the convolution operation, $\mathbf{F}$ is the column-wise version of the filter $\mathbf{f}$, i.e., $\mathbf{F} = [f_{1,1}, \ldots, f_{1,c}, \ldots, f_{c,1}, \ldots, f_{c,c}]^T \in \mathrm{R}^{cc \times 1}$ and $\mathbf{C} \in \mathrm{R}^{c^2 \times rl}$ is a block Toeplitz matrix formed by $\mathbf{m}$, such that the matrix $\mathbf{C}$ is constituted by the shifted and rearranged versions of the image $\mathbf{m}$. In addition, the operation Matrix returns the result of $\mathbf{F}^T \times \mathbf{C} \in \mathrm{R}^{1 \times rl}$ into the matrix format. Fig. 1, shows the steps of describing a two dimensional convolution by the matrix analysis, according to Ref. (Jayaraman et al., 2011).

A relevant criterion for determining $\mathbf{F}$ is energy, such that the filter $\mathbf{F}$ maximizing the energy of features is the appropriate filter for which we look. So, $\text{Var}(\mathbf{F}^T \times \mathbf{C})$ should be maximized where Var denotes the variance. As it is clear, the maximum of $\text{Var}(\mathbf{F}^T \times \mathbf{C})$ will not be achieved for finite $\mathbf{F}$, so a normalization constraint should be added to the maximization problem. One can easily propose the constraint $\mathbf{F}^T \times \mathbf{F} = 1$, that is, the sum of squares of elements of $\mathbf{F}$ equals 1. This constraint doesn't impose the filter $\mathbf{F}$ to be a high-pass, low-pass or band-pass filter. So, any type of the filter can be obtained by this constraint. The maximization problem is:

$$J(\mathbf{F_1}) = \arg\max_{\mathbf{F_1}} \{\text{Var}(\mathbf{F_1^T} \times \mathbf{C})\} - l_1(\mathbf{F_1^T} \times \mathbf{F_1} - 1) \quad (3)$$

where J is the objective function which should be maximized. This eigenvalue problem has to be solved to find the filter $\mathbf{F_1}$ i.e. the first filter extracting the first spatial feature. Differentiation of equation (3) with respect to $\mathbf{F_1}$ gives:

$$\mathbf{\Omega_c F_1} - l \mathbf{F_1} = 0 \quad (4)$$

The filter $\mathbf{F_1}$ is composed of the eigenvector of $\mathbf{\Omega_c} \in \mathrm{R}^{c^2 \times c^2}$ corresponding to its first largest eigenvalue $l_1$, where $\mathbf{\Omega_c}$ is the covariance matrix of $\mathbf{C}$. The second filter $\mathbf{F_2}$, should maximize $\text{Var}(\mathbf{F_2^T} \times \mathbf{C}) = \mathbf{F_2^T} \times \mathbf{\Omega_c} \times \mathbf{F_2}$ subject to being uncorrelated with $\mathbf{F_1}$, or equivalently subject to $\text{cov}[\mathbf{F_2^T} \times \mathbf{C}, \mathbf{F_1^T} \times \mathbf{C}] = 0$. So:

$$\text{Cov}[\mathbf{F_2^T} \times \mathbf{C}, \mathbf{F_1^T} \times \mathbf{C}] = \mathbf{F_2^T} \times \mathbf{\Omega_c} \times \mathbf{F_1} = $$
$$\mathbf{F_1^T} \times \mathbf{\Omega_c} \times \mathbf{F_2} = \mathbf{F_2^T} \times l_1 \mathbf{F_1} = l_1 \mathbf{F_2^T} \times \mathbf{F_1} = l_1 \mathbf{F_1^T} \times \mathbf{F_2} = 0 \quad (5)$$

Now, the optimization problem is:

$$J(\mathbf{F_2}) = \arg\max_{\mathbf{F_2}} \{\mathbf{F_2^T} \times \mathbf{\Omega_c} \times \mathbf{F_2}\} - l_2(\mathbf{F_2^T} \times \mathbf{F_2} - 1) - \Psi \mathbf{F_2^T} \times \mathbf{F_1} \quad (6)$$

where $l_2$ and $\Psi$ are the Lagrange multipliers. Differentiation with respect to $\mathbf{F_2}$ and then multiplication of the result on the left by $\mathbf{F_1^T}$ gives:

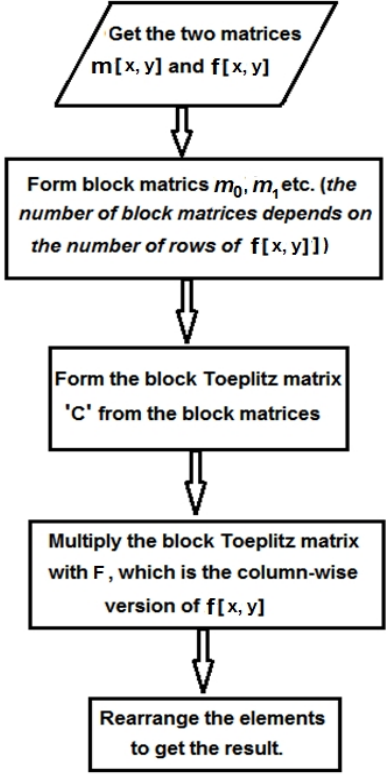

**Fig. 1**. Steps of a two dimensional convolution by the matrix analysis (Jayaraman et al., 2011).

$$\mathbf{F}_1^T \times \mathbf{\Omega}_c \times \mathbf{F}_2 - l_2 \mathbf{F}_1^T \times \mathbf{F}_2 - \Psi \mathbf{F}_1^T \times \mathbf{F}_1 = 0 \quad (7)$$

Because the first two terms are zero and $\mathbf{F}_1^T \times \mathbf{F}_1 = 1$, one can easily conclude $\Psi = 0$. So the optimization problem of equation (6) can be simplified to:

$$J(\mathbf{F}_2) = \arg\max_{\mathbf{F}_2} \{\mathbf{F}_2^T \times \mathbf{\Omega}_c \times \mathbf{F}_2\} - l_2(\mathbf{F}_2^T \times \mathbf{F}_2 - 1) \quad (8)$$

Similar to the solution of Eq. (3), $l_2$ is the second largest eigenvalue of $\mathbf{\Omega}_c$, and $\mathbf{F}_2$ is the corresponding eigenvector. It can be shown that $\mathbf{F}_3, \mathbf{F}_4, \ldots, \mathbf{F}_n$ are the eigenvectors of $\mathbf{\Omega}_c$ corresponding to $l_3, l_4, \ldots, l_n$, the third to $n^{th}$ largest eigenvalues. This solution is similar to the PCA usually used for extraction of the spectral features, but this time PCA is applied in the spatial domain. Note that the maximum number of orthogonal filters which can be designed is $c^2$ (because the dimension of $\mathbf{\Omega}_c$ is $c \times c$), and consequently, the maximum number of features corresponding to each pixel is equal to $c^2$.

Here several efficient and directional filters were designed, by which one can extract the spatial features of hyperspectral images and feed these features to the classifiers.

The proposed method has a reasonable computational complexity. It is just needed to compute a covariance matrix ($\mathbf{\Omega}_c$) and compute its eigen vectors to design the filter set.

## 3. Constructing the energy profile

In order to apply the filter set to hyperspectral data, computing the principal components of the hyperspectral image is the first step. The most significant principal components, which are the components with maximum energy, are used as base images for the energy profile, i.e., a profile based on more than one original image. When the proposed filter set is applied to hyperspectral data, a characteristic image needs to be extracted from the data. It was suggested to use the first principal component of the hyperspectral data for this purpose (Palmason et al., 2003). Because the principal component analysis is optimal for data representation in the mean square sense, this suggestion seems reasonable. But, it should be noted that with only one principal component, the hyperspectral data are reduced from potentially several hundred data channels into one dimensional data channel. Moreover, although the first principal component represents most of the variation in the data, other principal components contain some important information. Thus, similar to Ref. (Benediktsson et al., 2005), we make use of several different principal components to build an energy profile. In this paper, we make use of the principal components that account for around 90% of the total variation in the hyperspectral data. For example, if two principal components contain 90% of the variation, the energy profile would be a double profile where the first profile would be based on the first principal component and the second profile would be based on the second principal component (see Fig. 2). For each profile, the corresponding principal component is considered as an image to which the proposed filter set is applied. By the proposed filter set, the features of the principal components are extracted. These features constructing the energy profile are fed into an arbitrary classifier to obtain the classification results.

## 4. Experimental results

The goal of this section is to evaluate the performance of the proposed energy profile for classification of hyperspectral images. We apply the proposed method to the two well-known real hyperspectral data sets and classify these images by the energy profile. As mentioned before, the classifier used in our experiments is SVM (Huang et al., 2015; Guo and Boukira, 2015) with polynomial kernel of degree 3.

The SVM classifier is implemented using LIBSVM in which the default kernel parameter values i.e. $\gamma = 1/$Number of features, and $c_0 = 0$ are used (Chang and Lin, 2001). A number of labeled samples of each class are randomly selected to train the SVM classifier. In order to study the effect of the training sample size on the proposed feature extraction method, we have considered four different proportional schemes for the number of training samples: (a) 1%, (b) 5%, (c) 10% and (d) 12.5%. A point which should be considered is that there are some classes with small number of labeled samples in Indian Pines data. To ensure that we have enough training samples for the SVM classifier, a minimum of 3 samples per class is considered in schemes (a) and (b). In other words, we have $n_l = \max(3, 0.01 N_l)$ for scheme (a) and $n_l = \max(3, 0.05 N_l)$ for scheme (b), where $N_l$ is the number of labeled samples of the $l^{th}$ class. Although the size of parameter $c$ is dependent on the spatial structure and texture of the remotely sensed image, this parameter should be large enough to guarantee capturing the spatial information of the image. The size of $c$ is selected 35 in our experiments, because the improvement is not acceptable for the filter set with bigger size.

**Table 1. Sixteen categories and corresponding number of labeled pixels in the Indian Pines data**

| No | Category | pixels | No | Category | pixels |
|---|---|---|---|---|---|
| * | Alfalfa | 46 | * | Oats | 20 |
| * | Corn-no till | 1428 | * | Soybeans-no till | 972 |
| * | Corn-min till | 830 | * | Soybeans-min till | 2455 |
| * | Corn | 237 | * | Soybeans-clean | 593 |
| * | Grass/pasture | 483 | * | Wheat | 205 |
| * | Grass/trees | 730 | * | Woods | 1265 |
| * | Grass/pasture-mowed | 28 | * | Bldg-Grass-Tree-Drives | 386 |
| * | Hay-windowed | 478 | * | Stone-Steel-Towers | 93 |

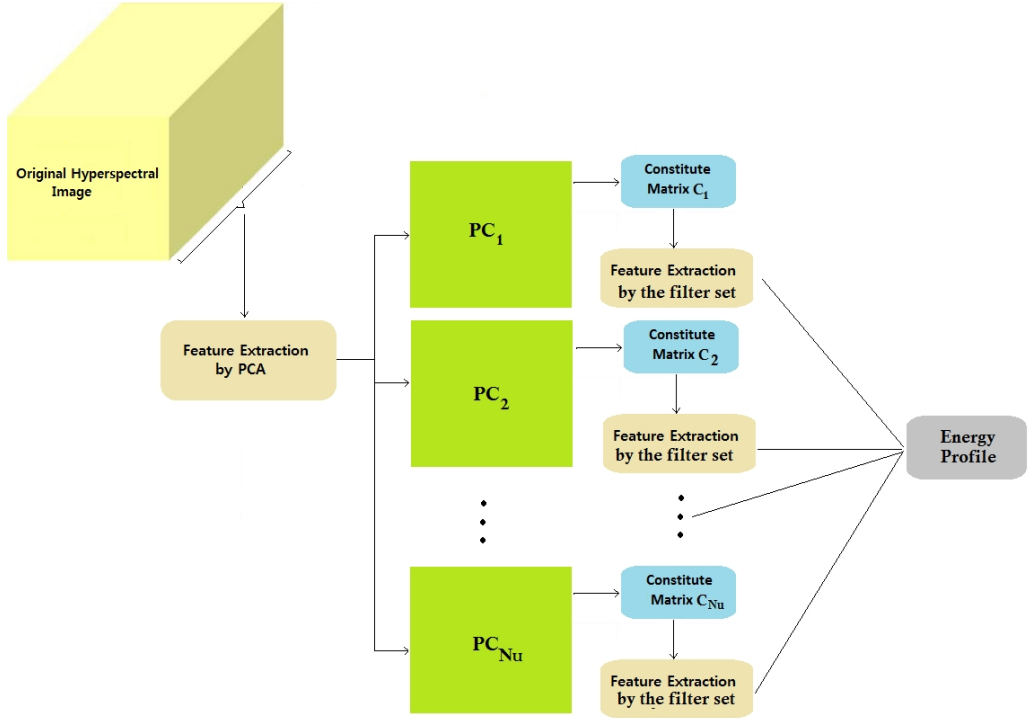

**Fig. 2.** Block-diagram of the proposed energy profile.

**Table 2.** OA, AA, and Kappa statistics ($k$) obtained by the SVM classifier on the Indian Pines data. Different combinations of the extracted features are used. The symbol '+' denotes stacking.

| | | Training set (randomly selected) | | | | | | | | | | | |
|---|---|---|---|---|---|---|---|---|---|---|---|---|---|
| | | Scheme (a) 1% | | | Scheme (b) 5% | | | Scheme (c) 10% | | | Scheme (d) 12.5% | | |
| Case | Input features to SVM | OA | AA | $k$ | OA | AA | $k$ | OA | AA | $k$ | OA | AA | $k$ |
| * | HS | 57.81 | 59.71 | 51.28 | 73.66 | 71.43 | 69.42 | 78.89 | 75.49 | 75.47 | 80.34 | 76.32 | 77.12 |
| * | HS+MP | 72.07 | 73.72 | 67.75 | 89.76 | 87.81 | 88.09 | 93.99 | 92.41 | 92.99 | 94.89 | 92.60 | 94.03 |
| * | HS+MP+Gab | 71.59 | 74.68 | 67.20 | 90.64 | 89.10 | 89.11 | 95.26 | 93.97 | 94.46 | 96.01 | 93.52 | 95.34 |
| * | HS+MP+Gab+SGl | 71.81 | 75.18 | 67.45 | 90.48 | 88.69 | 88.93 | 95.20 | 93.41 | 94.39 | 96.10 | 93.16 | 95.44 |
| * | HS+MP+Gab+SGl+Gl | 71.66 | 75.40 | 67.22 | 90.86 | 89.67 | 89.36 | 95.43 | 93.79 | 94.66 | 96.23 | 93.41 | 95.59 |
| * | PCA | 63.96 | 65.98 | 58.35 | 78.19 | 76.92 | 74.65 | 81.82 | 79.51 | 78.88 | 82.87 | 81.68 | 80.08 |
| * | PCA+MP | 77.06 | 76.95 | 73.49 | 92.39 | 90.52 | 91.14 | 95.68 | 93.49 | 94.95 | 95.87 | 93.25 | 95.17 |
| * | PCA+MP+Gab | 73.39 | 77.10 | 69.35 | 92.45 | 91.07 | 91.21 | 96.11 | 94.38 | 95.45 | 96.67 | 94.90 | 96.11 |
| * | PCA+MP+Gab+SGl | 72.33 | 75.29 | 68.14 | 91.56 | 89.57 | 90.17 | 95.45 | 92.95 | 94.69 | 96.10 | 94.48 | 95.44 |
| * | PCA+MP+Gab+SGl+Gl | 70.38 | 74.43 | 65.87 | 91.41 | 89.65 | 90.00 | 95.67 | 93.47 | 94.94 | 96.41 | 94.57 | 95.80 |
| * | LDA | 64.52 | 67.96 | 58.67 | 76.79 | 76.86 | 73.06 | 81.11 | 76.72 | 78.06 | 82.41 | 79.47 | 79.53 |
| * | LDA +MP | **79.19** | **78.96** | **75.94** | 92.97 | 90.51 | 91.81 | 95.45 | 92.77 | 94.69 | 96.11 | 93.20 | 95.45 |
| * | LDA +MP+Gab | 74.58 | 77.51 | 70.69 | 92.54 | 90.49 | 91.31 | 96.05 | 93.46 | 95.38 | 96.97 | 94.44 | 96.45 |
| * | LDA +MP+Gab+SGl | 72.30 | 75.99 | 68.10 | 91.63 | 89.25 | 90.24 | 95.44 | 92.73 | 94.67 | 96.31 | 93.22 | 95.68 |
| * | LDA+MP+Gab+SGl+Gl | 71.84 | 76.37 | 67.56 | 91.77 | 89.84 | 90.41 | 95.82 | 93.20 | 95.12 | 96.48 | 94.02 | 95.88 |
| * | Proposed | 73.67 | 75.72 | 69.31 | **94.54** | **94.15** | **93.62** | **97.78** | **96.95** | **97.40** | **98.36** | **97.56** | **98.08** |

The first hyperspectral data set is Indian Pines one, acquired on 12 June 1992 by the AVIRIS sensor, covering a 2.9 km × 2.9 km portion of northwest Tippecanoe County, Indiana, USA. Two-thirds of this scene is agriculture, and one-third of it, is forest or other natural perennial vegetation. This data includes 220 bands with 145 × 145 pixels and a spatial resolution of 20 m. Twenty bands of this data set have been removed because they cover the water absorption spectrum band (104–108, 150–163, 220). The corrected data set including 200 bands has been used in our experiments. There are 16 different land-cover classes available in the original ground-truth map. Table 1 lists the number of labeled pixels of each class.

Due to the presence of mixed pixels in all available classes and because there are classes with small number of labeled pixels, this data set constitutes a challenging classification problem (Song et al., 2014). Several combinations of spectral and spatial features (16 cases) which are very successful according to (Mirzapour and Ghassemian, 2015) are fed into the SVM classifier and the classification results are given in table 2. Note that the results of all the tables (tables 2, 4 and 5) are reported after running 50 times (the Monte Carlo method) and averaging the values. The abbreviations in these tables are: OA is overall accuracy, AA is average accuracy, $k$ is kappa statistics, HS is hyperspectral image, MP is Morphological Profile, Gab is Gabor features, Gl is GLCM (gray-level co-occurrence matrix) features and SGl is segmentation-based GLCM features. In addition, Fig. 3 shows the sample classification map obtained by the proposed method.

**Table 3. Sixteen categories and corresponding number of labeled pixels in the Indian Pines data**

| No | Category | pixels | No | Category | pixels |
|---|---|---|---|---|---|
| * | Brocoli_green_weeds_1 | 2009 | * | Soil_vinyard_develop | 6203 |
| * | Brocoli_green_weeds_2 | 3726 | * | Corn_senesced_weeds | 3278 |
| * | Fallow | 1976 | * | Lettuce_romaine_4wk | 1068 |
| * | Fallow_rough_plow | 1394 | * | Lettuce_romaine_5wk | 1927 |
| * | Fallow_smooth | 2678 | * | Lettuce_romaine_6wk | 916 |
| * | Stubble | 3959 | * | Lettuce_romaine_7wk | 1070 |
| * | Celery | 3579 | * | Vinyard_untrained | 7268 |
| * | Grapes_untrained | 11271 | * | Vinyard_vertical_trellis | 1807 |

The second data set is Salinas one, collected by the AVIRIS sensor over Salinas Valley, California, USA. Each spectral band has $512 \times 217$ pixels, where the geometric resolution of each pixel is 3.7 m. As with the Indian Pines scene, the 20 water absorption bands (108–112, 154–167, 224) are discarded, resulting in a corrected image containing 204 spectral bands. This data includes 16 agricultural land covers with very similar spectral signatures (Plaza et al., 2005). The number of labeled pixels of each class is listed in table 3. The mentioned indices are shown in table 4 according to the corresponding schemes. Moreover, Fig. 4 shows the sample classification map obtained by the proposed method.

At the end of this section, we provide a comparative assessment. The proposed method is compared with some recently proposed spectral–spatial feature extraction methods. To do a reliable comparison, the classification results for SVM-CK (Camps-Valls et al., 2006) are obtained by using the publicly available codes. But, the results for other methods are reported from the corresponding papers. Thus, some results are absent. The results of this comparison is reported in table 5. As can be seen from this table, finding a unique feature extraction method which gives the best results for the different number of training samples and for the different data sets, is difficult. In fact, it is not exaggerated if one say that there does not exist such features (Kuo et al. 2009). As can be seen from table 5, the proposed feature extraction method is almost certainly the best when there are enough training samples. In addition, the proposed method has a satisfactory performance when the number of training samples is low.

## 5. Conclusion

A new spatial-spectral feature extraction method for classification of hyperspectral images was proposed. By maximizing the energy of the extracted features, and by uncorrelating the filters, coefficients of the spatial filters are designed. To provide the energy profile including spectral-spatial features of hyperspectral data, the PCA is applied to the hyperspectral data at the first step. Then, the proposed filter set is applied to the principal components that account for around 90% of the total variation in the hyperspectral data. To evaluate the performance of the proposed feature extraction method, the SVM classifier with the polynomial kernel of degree 3 was used to classify the extracted features. Two real hyperspectral data sets namely, Indian Pines and Salinas data were used in our experiments. The experimental results have demonstrated that the proposed feature extraction method can provide efficient features for the classification application.

**Table 4. OA, AA, and Kappa statistics ($k$) obtained by the SVM classifier on the Salinas data. Different combinations of the extracted features are used.**

| | | Training set (randomly selected) | | | | | | | | | | | |
|---|---|---|---|---|---|---|---|---|---|---|---|---|---|
| | | Scheme (a) 1% | | | Scheme (b) 5% | | | Scheme (c) 10% | | | Scheme (d) 12.5% | | |
| Case | Input features to SVM | OA | AA | $k$ | OA | AA | $k$ | OA | AA | $k$ | OA | AA | $k$ |
| * | HS | 88.01 | 92.88 | 86.60 | 90.83 | 95.76 | 89.75 | 91.99 | 96.49 | 91.05 | 92.18 | 96.54 | 91.25 |
| * | HS+MP | 95.15 | 96.51 | 94.58 | 97.88 | 98.68 | 97.63 | 98.36 | 99.05 | 98.17 | 98.57 | 99.17 | 98.40 |
| * | HS+MP+Gab | 95.08 | 96.58 | 94.50 | 98.22 | 98.88 | 98.01 | 98.83 | 99.32 | 98.70 | 99.03 | 99.41 | 98.91 |
| * | HS+MP+Gab+SGl | 95.76 | 96.90 | 95.26 | 98.42 | 98.86 | 98.24 | 99.10 | 99.38 | 99.00 | 99.21 | 99.45 | 99.12 |
| * | HS+MP+Gab+SGl+Gl | **95.82** | **96.95** | **95.33** | 98.57 | 98.95 | 98.40 | 99.24 | 99.44 | 99.15 | 99.38 | 99.52 | 99.31 |
| * | PCA | 87.72 | 92.27 | 86.26 | 90.13 | 94.63 | 88.95 | 90.09 | 95.05 | 88.92 | 90.62 | 95.34 | 89.50 |
| * | PCA+MP | 93.78 | 94.99 | 93.04 | 97.52 | 98.02 | 97.23 | 98.32 | 98.77 | 98.12 | 98.60 | 99.00 | 98.43 |
| * | PCA+MP+Gab | 94.01 | 95.37 | 93.30 | 98.03 | 98.53 | 97.80 | 98.82 | 99.19 | 98.67 | 99.03 | 99.39 | 98.91 |
| * | PCA+MP+Gab+SGl | 95.12 | 96.20 | 94.55 | 98.23 | 98.67 | 98.02 | 99.01 | 99.25 | 98.89 | 99.19 | 99.42 | 99.09 |
| * | PCA+MP+Gab+SGl+Gl | 94.86 | 96.04 | 94.26 | 98.36 | 98.73 | 98.17 | 99.01 | 99.26 | 98.89 | 99.23 | 99.44 | 99.14 |
| * | LDA | 89.18 | 94.05 | 87.91 | 92.21 | 96.14 | 91.29 | 93.03 | 96.68 | 92.19 | 93.25 | 96.83 | 92.45 |
| * | LDA +MP | 93.88 | 95.32 | 93.16 | 97.65 | 98.18 | 97.37 | 98.42 | 98.77 | 98.23 | 98.58 | 98.89 | 98.41 |
| * | LDA +MP+Gab | 94.18 | 95.66 | 93.49 | 98.00 | 98.53 | 97.77 | 98.86 | 99.23 | 98.73 | 99.04 | 99.35 | 98.93 |
| * | LDA +MP+Gab+SGl | 95.30 | 96.47 | 94.75 | 98.26 | 98.70 | 98.06 | 99.01 | 99.23 | 98.90 | 99.25 | 99.44 | 99.16 |
| * | LDA+MP+Gab+SGl+Gl | 95.11 | 96.34 | 94.53 | 98.31 | 98.72 | 98.11 | 99.06 | 99.29 | 98.95 | 99.27 | 99.45 | 99.19 |
| * | Proposed | 93.89 | 95.36 | 93.08 | **99.10** | **99.50** | **98.99** | **99.58** | **99.80** | **99.53** | **99.62** | **99.84** | **99.58** |

**Table 5. Overall accuracy of the proposed method compared with some recently proposed classification methods**

| Data set | Number of Training samples | Diverse AdaBoost SVM† | FODPSO + MSS (R = 5) + SVM‡ | EMAP (KPCA)+RF classifier€ | SUnSAL$_{EMAP}$¶ | SVM CK" | Proposed |
|---|---|---|---|---|---|---|---|
| Indian Pines | 1% | - | - | - | - | **73.7** | **73.7** |
| | 5% | - | - | 88.7 | **95.0** | 91.4 | 94.5 |
| | 10% | 92.6 | - | 92.7 | 96.8 | 94.9 | **97.8** |
| | 12.5% | - | - | - | - | 95.8 | **98.4** |
| Salinas | 1% | - | - | - | - | **95.6** | 93.9 |
| | 5% | - | - | - | - | 98.7 | **99.1** |
| | 10% | 94.9 | - | - | - | 98.9 | **99.6** |
| | 12.5% | - | 99.1 | - | - | 99.1 | **99.6** |

†Reported from Ref. (Ramzi et al., 2013). ‡ Reported from Ref. (Ghamisi et al., 2014). €Reported from Ref. (Bernabé et al., 2014). ¶ Reported from Ref. (Sonf et al., 2014). " Reported from Ref. (Camps-Valls et al., 2006).

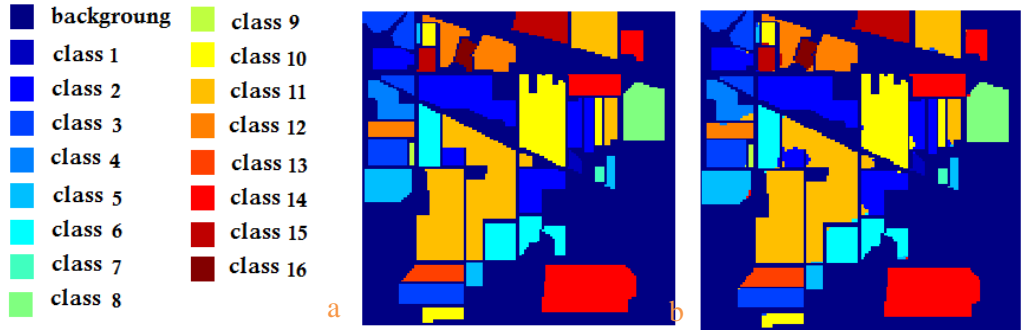

**Fig. 3.** The classification result for the AVIRIS Indian Pines scene. a) Ground truth map. b) The classification result of the proposed method (12.5% of the available labeled data is used).

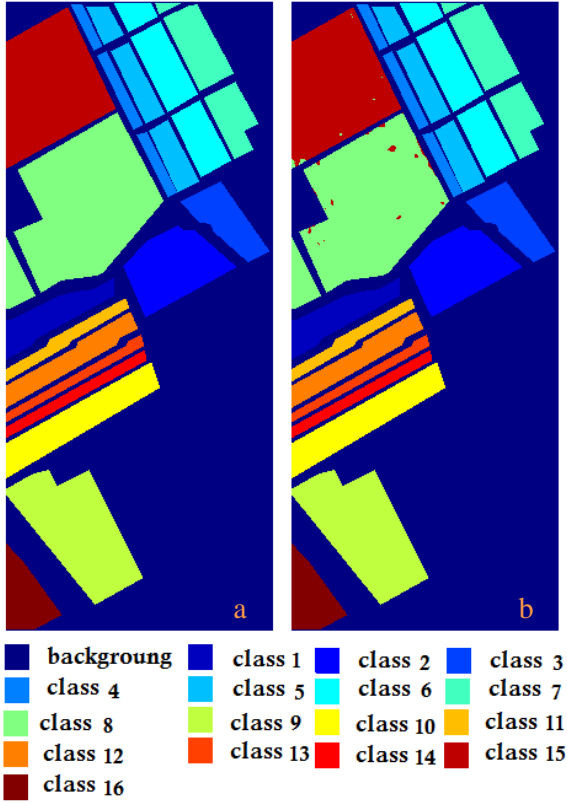

**Fig. 3.** The classification result for the Salinas scene. a) Ground truth map. b) The classification result of the proposed method (12.5% of the available labeled data is used).